\documentclass[letterpaper, 10 pt, conference]{ieeeconf}
\IEEEoverridecommandlockouts
\usepackage{cite}
\usepackage{amsmath,amssymb,amsfonts}
\usepackage{algorithmic}
\usepackage[xetex]{graphicx}
\usepackage{textcomp}
\usepackage{todonotes}
\usepackage{xcolor}
\def\BibTeX{{\rm B\kern-.05em{\sc i\kern-.025em b}\kern-.08em
    T\kern-.1667em\lower.7ex\hbox{E}\kern-.125emX}}

\usepackage[acronym, toc]{glossaries}
\usepackage{siunitx}
\usepackage[pdftex,bookmarks=true]{hyperref}
\usepackage[pdftex]{hyperref}
\usepackage[noabbrev]{cleveref}
\usepackage{subcaption}
\usepackage{booktabs}

\newacronym{drl}{DRL}{Deep Reinforcement Learning}
\newacronym{rl}{RL}{Reinforcement Learning}
\newacronym{mdp}{MDP}{Markov Decision Process}
\newacronym{ann}{ANN}{Artificial Neural Network}
\newacronym{ddpg}{DDPG}{Deep Deterministic Policy Gradient}
\newacronym{ros}{ROS}{Robot Operating System}
\newacronym{her}{HER}{Hindsight Experience Replay}
\newacronym{sgd}{SGD}{Stochastic Gradient Descent}
\newacronym{shap}{SHAP}{SHapley Additive exPlanations}
\newacronym{xai}{XAI}{eXplainable Artificial Intelligence}
\newacronym{urdf}{URDF}{Unified Robot Description Format}
\newacronym{dof}{DOFs}{degrees-of-freedom}
\newacronym{rnn}{RNN}{Recurrent Neural Network}
\newacronym{dqn}{DQN}{Deep Q-network}
\newacronym{relu}{ReLU}{Rectified Linear Unit}
\newacronym{ppo}{PPO}{Proximal Policy Optimization}
\newacronym{sac}{SAC}{Soft Actor-Critic}
\newacronym{ai}{AI}{Artificial Intelligence}
\newacronym{lime}{LIME}{Local Interpretable Model-agnostic Explanations}
\newacronym{ml}{ML}{Machine Learning}
\newacronym{td3}{TD3}{Twin Delayed Deep Deterministic Policy Gradient}

\begin{document}
\title{\LARGE \bf
Robotic Lever Manipulation using Hindsight Experience Replay and Shapley Additive Explanations
}

\author{Sindre Benjamin Remman$^{1}$ and Anastasios M. Lekkas$^{2}$
\thanks{$^{1}$Sindre Benjamin Remman is with Department of Engineering Cybernetics, 
        Norwegian University of Science and Technology (NTNU), Trondheim, Norway
        {\tt\small sindre.b.remman@ntnu.no}}%
\thanks{$^{2}$Anastasios M. Lekkas is with Department of Engineering Cybernetics,
        Centre for Autonomous Marine Operations and Systems (AMOS),
        Norwegian University of Science and Technology (NTNU), Trondheim, Norway
        {\tt\small anastasios.lekkas@ntnu.no}}%
}

\maketitle

\begin{abstract}
This paper deals with robotic lever control using Explainable Deep Reinforcement Learning. First, we train a policy by using the Deep Deterministic Policy Gradient algorithm and the Hindsight Experience Replay technique, where the goal is to control a robotic manipulator to manipulate a lever. This enables us both to use continuous states and actions and to learn with sparse rewards. Being able to learn from sparse rewards is especially desirable for Deep Reinforcement Learning because designing a reward function for complex tasks such as this is challenging. We first train in the PyBullet simulator, which accelerates the training procedure, but is not accurate on this task compared to the real-world environment. After completing the training in PyBullet, we further train in the Gazebo simulator, which runs more slowly than PyBullet, but is more accurate on this task. We then transfer the policy to the real-world environment, where it achieves comparable performance to the simulated environments for most episodes. To explain the decisions of the policy we use the SHAP method to create an explanation model based on the episodes done in the real-world environment. This gives us some results that agree with intuition, and some that do not. We also question whether the independence assumption made when approximating the SHAP values influences the accuracy of these values for a system such as this, where there are some correlations between the states.
\end{abstract}
\glsresetall

\section{Introduction}

\gls{drl}, which is the fusion of traditional \gls{rl} and \glspl{ann}, has since the early 2010s been used to solve a variety of difficult problems. The first successful application of \gls{drl} was done in \cite{mnih2013playing}, where the authors made a \gls{drl} algorithm, called \gls{dqn}, that could learn how to play Atari 2600 video games directly from high-dimensional screen pixel input. After that, \gls{drl} has shown promise in various fields: \gls{drl} was used to create AlphaZero, which beat world-champion computer programs at Chess, Shogi, and Go \cite{AlphaZero}; Agent57 was created in 2020, the first \gls{drl} agent that learned to play all 57 Atari 2600 games in the OpenAI gym with super-human performance \cite{badia_agent57_2020}; \gls{drl} has also made great strides in the field of robotics \cite{DDPGpaper, haarnoja2018softOff-Policy, andrychowicz2017hindsight_her, levine2015endtoend, peng2017simtoreal}.

One commonly used \gls{drl} algorithm is \gls{ppo}, which was presented in \cite{PPO_paper}. \gls{ppo} has been used in various tasks such as: quadrotor control \cite{lopes_intelligent_2018}, electrical vehicle charging with uncertain wind power \cite{EV_charging_uncertain_wind_power}, and for developing a step climbing method for a crawler type rescue robot \cite{step_climbing_method}. A problem with \gls{ppo} is that it has a poor sample efficiency. This is because it is an \textit{on-policy} algorithm, which means that it cannot reuse past experiences, and needs to collect new samples every time it optimizes its parameters \cite{haarnoja2018softOff-Policy}. The algorithm used in this paper is the \gls{ddpg} algorithm, presented in \cite{DDPGpaper}, which unlike \gls{ppo}, is an \textit{off-policy} algorithm. This means that it can reuse past experiences. \gls{ddpg} has been used in tasks such as: traffic signal control \cite{traffic_signal_ddpg}, control of unmanned surface vehicles \cite{Wang_unmanned_surface_vehicle_ddpg}, and flight control \cite{flight_control_ddpg}. A disadvantage with \gls{ddpg} is that it is sensitive with regards to hyperparameters, which can require significant effort when tuning the algorithm \cite{haarnoja2018softOff-Policy}. \gls{ppo} and \gls{ddpg} use a stochastic and deterministic policy respectively. Stochastic policies are appropriate for an agent that needs to adapt to stochasticity in the environment, but for this paper, deterministic policies are more suitable. This is because a deterministic policy's decisions arguably will be easier to explain. 

Even though \gls{drl} shows great promise, and has been used to solve remarkably different problems, there are still many drawbacks with \gls{drl}: it is hard to do intelligent exploration, instead of just random exploration; training a policy can take a significant amount of time; safety guarantees during safety-critical operations are still lacking; it can be difficult and time-consuming to design a good reward function that does not have any unwanted side-effects; and it can be difficult to understand how a \gls{drl} policy makes its decision, which ties into whether or not the policy can be trusted. The last drawback mentioned here is mainly because of the black-box nature of \glspl{ann}. 

Based on the recent advances in \gls{ml}, researchers are now looking at how to explain the decisions made by \gls{ml} agents. Many of the most prominent applications of \gls{ml} are now done using models that have a black-box nature. This makes it especially demanding to understand how the agents make their decisions. The field that handles these types of explanation problems is collectively called \gls{xai} and has garnered increasing interest these last years \cite{Barredo_Arrieta_2020, belle_principles_2020}. \gls{lime} is one of the more popular \gls{xai} methods, which learns an interpretable model locally around the prediction of the \gls{ml} model \cite{ribeiro2016whyshoulditrustyou_lime}. Another \gls{xai} method is the \textit{Integrated Gradients} method, which integrates the gradients of an \gls{ann} on a straight-line path from a baseline $x'$ to an input $x$ to explain how the inputs to an \gls{ann} affect its prediction \cite{sundararajan_axiomatic_2017_integrated_gradients}. The \gls{xai} method used in this paper is called \gls{shap} \cite{NIPS2017_7062}. Similar to \gls{lime}, \gls{shap} is an additive feature attribution method. \gls{shap} has some desirable properties that \gls{lime} does not have, but \gls{lime} can generally generate explanations faster. \gls{shap} has also been shown empirically to give explanations that better agree with human intuition \cite{NIPS2017_7062}.

In this paper, we examine how to alleviate the last two of the drawbacks mentioned above by using the \gls{rl} technique \gls{her} \cite{andrychowicz2017hindsight_her} and \gls{shap}, where the objective is to manipulate a lever using a robotic manipulator. \gls{her} enables the usage of a simple reward function for a complex system such as a robotic manipulator, and \gls{shap} provides an approximation of the contribution of each state to each action. The results of this paper are an extension of the first author's Master's thesis \cite{remman_thesis_2020}. We do these experiments in a platform that involves lever control with a robotic manipulator, more specifically the contributions of this paper are:

\begin{itemize}
    \item The design of an experimental setup including the OpenMANIPULATOR-X by ROBOTIS, which involves lever angle control and can correspond to many problems in the physical world. In addition to the real-world experimental setup, corresponding simulated environments including a model of the lever was created. We trained the policy in two simulated environments. After training, the policy was implemented on the physical manipulator.
    \item We customize the sparse reward function to the lever manipulation problem and train the simulated system using \gls{ddpg} and \gls{her}. 
    \item We implement \gls{shap} to understand how the actions are selected by the policy. In this way, we provide insight and can figure out if the system has learned behavior that conforms to human intuition. 
    
\end{itemize}
A related approach can be seen in \cite{wang_attribution-based_nodate}, where the authors also apply \gls{shap} to a \gls{drl} problem and then investigate how different background data influence the explanations. However, this is done on a much less complicated system than our system, which only has four states and one action, where we have 20 states and four actions. Another related approach is shown in \cite{robot_failure_mode_prediction}, where they use \gls{ml} to predict failure modes in robot grasping, and compare how these failures can be explained using both inherently interpretable \gls{ml} models and black-box \gls{ml} models. Unlike our work, they do this solely in a simulated environment, where we use a real-world environment in addition to simulated environments.

The remaining sections in this paper are organized as follows: the Preliminaries, where we discuss some of the necessary background theory; the Methodology, where we discuss the task and experimental setup; the Results and Discussion, where we show and discuss the results from the lever-manipulation task performed in this paper; and finally the Conclusion, where we give a brief overview of the main points from this paper. 


\section{Preliminaries}
This section briefly examines and explains the necessary theory for the rest of the paper. Firstly, we give the fundamental concepts in \gls{rl} and \gls{drl}. Secondly, the \gls{her} technique is motivated and explained. Lastly, the theory and properties of \gls{shap} are examined.
\subsection{Reinforcement learning}
In \gls{rl}, we assume that we can model the environment as a \gls{mdp}. An \gls{mdp} is defined as a tuple $<\mathcal{S}, \mathcal{A}, T, R>$, where $\mathcal{S}$ is a state-space, $\mathcal{A}$ is an action-space, $T$ is a Markovian transition model, and $R$ is a reward function. A Markovian transition model is a proper probability distribution over the next possible states given the current state and the action taken in the current state. A Markovian transition model satisfies the Markov property, which means that only the current state and action are relevant for the distribution of the next state \cite[p.11]{ReinfLearnState}. The reward function returns a scalar number, the reward, based on the transition $R(s_t,a_t,s_{t+1})=r_t$.

The goal for an \gls{rl} agent is to find the optimal policy $\pi^*$, which is the policy that maximizes the long-term expected reward, here defined by the discounted infinite horizon model:
\setlength{\textfloatsep}{2pt}
\begin{equation}
    E[\sum^\infty_{t=0}\gamma^t r_t],
\end{equation}
\setlength{\textfloatsep}{2pt}
where $\gamma \in [0,1]$ is a hyperparameter called the discount factor \cite[pp.13-15]{ReinfLearnState}.

The \gls{ddpg} algorithm used in this paper is an off-policy actor-critic algorithm that can tackle problems with continuous state- and action-spaces. As briefly mentioned in the introduction, off-policy method can use samples generated any time during training for optimization \cite{spinning_up_part_2}. Being actor-critic, \gls{ddpg} trains two \glspl{ann}. The first is the \emph{actor-network}, which serves the role of the policy, that is, it takes in the state of the environment and outputs what the action should be (equivalent to a controller). The second is the \emph{critic-network}, which is used to approximate the value of taking a certain action in a certain state (called a Q-value). The critic's role is to evaluate the performance of the actor. This evaluation is subsequently used for optimizing the actor's parameters. There were two crucial components for \gls{drl} that were introduced in \cite{mnih2013playing} and \cite{mnih_human-level_2015} in the context of \gls{dqn}. Both of these components are also used in \gls{ddpg}. The first component is the \emph{experience replay}, which ensures that training samples are independent, a requirement when training \glspl{ann}. This requirement is crucial to avoid overfitting because of the strong correlation between sequential samples \cite{mnih2013playing}. Since most optimization algorithms assume that samples are independent of each other, the training might be unsuccessful if samples are used for optimization in just the order they appear. Experience replay is used as a buffer that all transitions $\{s_t, a_t, r_t, s_{t+1}\}$ are stored in and recalled randomly from during \gls{ann} training. This means that the transitions are independent of each other, and that previous transitions are not forgotten and can be used multiple times. Experience replay buffers can only be used by off-policy algorithms \cite{mnih2013playing}. The second component introduced in \cite{mnih2013playing} and \cite{mnih_human-level_2015} is the \emph{target network}, which are copies of the original networks that are, for instance, gradually changed towards the parameters of the original networks (Polyak averaging), or copied from the original networks at a certain interval. Having a target that is changing fast becomes similar to trying to catch up to a moving target, so by using target networks the correlation with the target is decreased, which improves the training stability \cite{mnih_human-level_2015, DDPGpaper}.

We would like to note that there exist improvements over the \gls{ddpg} algorithm such as the state-of-the-art \gls{td3} algorithm presented in \cite{TD3_fujimoto_addressing_2018}. However, in \gls{rl}, there are no guarantees that any algorithm will work, and \gls{ddpg} has worked well for this system from the start. This choice should also not affect the later focus on explainability. 

We could also have included comparisons with other learning-based control methods for robotic manipulators. However, in the context of this paper, it is not in the scope to do this. This paper aims to study the sequence from the simulator to the real-world to explainability.

\subsection{Hindsight experience replay}\glsreset{her}
Arguably, one of the hardest parts of implementing \gls{rl} and \gls{drl} concerns how to engineer a good reward function. Give the reward too sparsely, and the agent will learn slowly, or not at all; give the reward too frequently, and the behavior of the agent may already be specified by the reward function. 

A novel technique that can enable \gls{rl} agents to learn from sparse rewards was presented in \cite{andrychowicz2017hindsight_her}. The approach is called \gls{her}, where the idea is to substitute the actual goals with virtual goals, which represent the goal-state that was actually achieved by the agent \cite{her_ingredients}. Consider an agent that has achieved a trajectory of $s_1,\dots, s_T$, but the goal state is not in this trajectory. In this case, with sparse rewards and ordinary \gls{drl}, the agent would receive the same reward for each step of the trajectory, and would not have any useful feedback for optimization. Even though the agent has not discovered how to reach the goal state, from another point of view, it has discovered how it can reach every other state in the trajectory. When using \gls{her}, any of the states within the trajectory can then be used as substituted goals, where the agent gets the same reward as it would if it achieved the real goal. In this way, the agent will get feedback which it can then use to optimize its parameters.

In this paper, the strategy that is used for selecting which state that should eligible for being substituted goals is called "future". For every transition stored in the experience replay, $k$ new versions of the transition are also stored, where the goal states are substituted with randomly selected achieved goal states that came from the same episode, but were observed after the transition. 

When using \gls{her}, it is useful to consider the state-space as consisting of two parts. The first part is the observation of the environment, the second part is the goal states. This way, when substituting goals, only the part of the state that contains the goal states needs to be substituted with the actual achieved goals.

\subsection{Shapley Additive Explanations}
\gls{shap} is a method developed in \cite{NIPS2017_7062}, followed by a library in python\footnote{\url{https://github.com/slundberg/shap}} that can help to identify which inputs are most important for a function's output. \gls{shap} is a model-agnostic and post-hoc \gls{xai} method, which means that it can work with any type of model or function, and it can explain decisions based on already generated data. It does this by approximating the Shapley values \cite{shapley_value_1952} for the input compared to the output. The Shapley values explain how each input contributes to the magnitude of the output. \gls{shap} is an additive feature attribution method, which means that it is a linear function of binary variables, of the form: 
\setlength{\textfloatsep}{2pt}
\begin{equation}\label{eq:additive_feature_attribution_method_form}
    g(z') = \phi_0 + \sum_{i=1}^M \phi_i z'_i,
\end{equation}
\setlength{\textfloatsep}{2pt}
"where $z' \in \{0,1\}$, $M$ is the number of simplified input features, and $\phi_i \in \mathbb{R}$" \cite{NIPS2017_7062}.
It can be shown that additive feature attribution methods have a single unique solution which has three desired properties \cite{NIPS2017_7062}:
\begin{itemize}
    \item Local accuracy
    \begin{itemize}
        \item When using an explanation model to approximate another model for a specific input $x$, the explanation model's output will match the other model's output for the simplified input $x'$ which corresponds to the input $x$.  
    \end{itemize}
    \item Missingness
    \begin{itemize}
        \item Requires features missing from the original input $x$ to have no impact: $$x'_i = 0 \rightarrow \phi_i = 0.$$
    \end{itemize}
    \item Consistency
    \begin{itemize}
        \item If a model changes so that a simplified input's contribution increases or stays the same, then that input's impact should not decrease.\\
    \end{itemize}
\end{itemize}

As stated above, there is only one solution to the explanation model  \eqref{eq:additive_feature_attribution_method_form} that satisfies the three properties just stated. This solution corresponds to the Shapley values \cite{NIPS2017_7062}:

\begin{equation}\label{eq:shap_equation}
    \phi_i(f,x) \hspace{-0.045cm} = \hspace{-0.045cm} \sum_{z'\subseteq x'}\frac{|z'|!(M-|z'|-1)!}{M!} [f_x(z')-f_x(z'\backslash i)],
\end{equation}

where $|z'|$ is the number of non-zero entries in $z'$, $z'\subseteq x'$ represents all $z'$ vectors where the non-zero entries are a subset of the non-zero entries of the simplified inputs $x'$, $f_x(z') = f(h_x(z'))$, $f$ is the original prediction model, and $h_x$ is a mapping function that maps simplified inputs to the original inputs \cite{NIPS2017_7062}. \Cref{eq:shap_equation} is a difficult equation to solve, as it relies on $f_x(z'\backslash i)$, which is the model that is going to be explained, without feature $i$ present. This is not straight-forward to achieve, for instance, for an \gls{ann}, because we would have to train several new models without the feature(s) present. The \gls{shap} library provides several methods to address these challenges and approximate these values. Some of these methods take advantage of a model's structure to more efficiently approximate the Shapley values. The \gls{shap} method that is used in this paper is called \emph{Deep SHAP} and is specific for \glspl{ann}.

\section{Methodology}
For the experiments done in this paper, we used both a physical and two simulated versions of the OpenMANIPULATOR-X robotic manipulator by ROBOTIS\footnote{\url{https://emanual.robotis.com/docs/en/platform/openmanipulator_x/overview/}}. Two simulators were used, Gazebo and PyBullet. The manipulator has 5 \gls{dof}, four for the joints of the manipulator, and one for the gripper. The physical robot and its Gazebo counterpart were controlled using the \gls{ros}, with packages developed by ROBOTIS\footnote{\url{https://github.com/ROBOTIS-GIT/open_manipulator}}. Also, we developed an environment in the PyBullet simulator to accelerate training, as will be explained in \Cref{sec:training_procedure}. The OpenAI Gym framework was used to create the \gls{rl} environments. The \gls{ddpg}+\gls{her} implementation used the PyTorch deep learning framework and was adapted from a GitHub repository created by A. Imran\footnote{\url{https://github.com/alishbaimran/Robotics-DDPG-HER}}.

\subsection{Lever manipulation task}
The goal for each episode in this task is to move the lever to a randomly selected goal angle. The starting position and goal position of the lever are randomly selected according to 
$$\theta_{start},\theta_{goal} \in \mathbb{R}: \theta_{start},\theta_{goal} \in [-1.0\text{ } \si{\radian}, 1.0\text{ } \si{\radian}],$$ with the additional constraint that $|\theta_{start}-\theta_{goal}|>0.4  \text{ }\si{\radian}.$

All policies in this paper are trained using \gls{her} and because of this, the reward function can be designed to only give sparse rewards on this task. The reward is given according to 
$$r =  \begin{cases}
    -1, & \text{if } |\theta_{lever}-\theta_{goal}|\geq0.025 \text{ }\si{\radian}\\
    0, & \text{if } |\theta_{lever}-\theta_{goal}|<0.025 \text{ }\si{\radian}
\end{cases},$$

where $0.025 \text{ }\si{\radian}$ is approximately $1.43^{\circ}$, which we deemed to be a sufficient accuracy for a task such as this.

\subsection{Experimental design}
\begin{figure}
\vspace{0.2cm}
\begin{subfigure}{.24\textwidth}
  \centering
  \includegraphics[width=\linewidth, height=3.3cm]{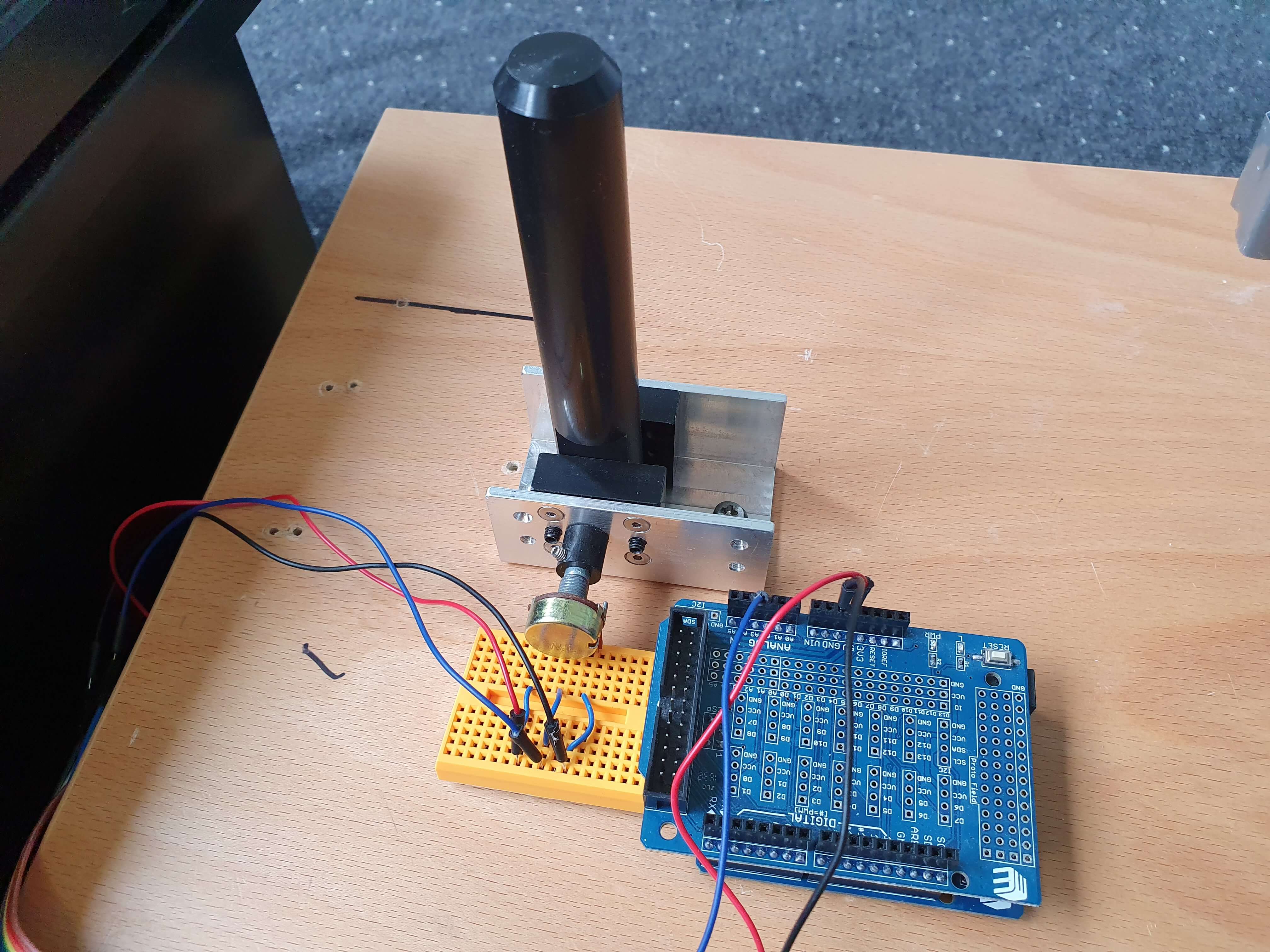}
  \caption{}
  \label{fig:lever_real}
\end{subfigure}
\begin{subfigure}{.24\textwidth}
  \centering
  \includegraphics[trim=0 3.55cm 0 2cm,clip,width=\linewidth,height = 3.3cm]{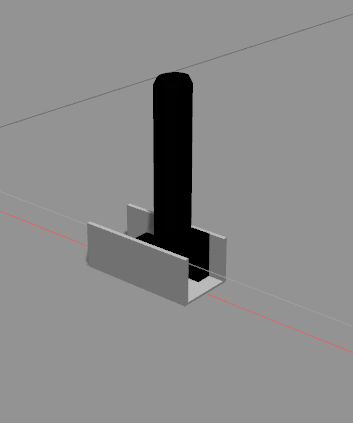}
  \caption{}
  \label{fig:lever_gazebo}
\end{subfigure}%
\caption{\textbf{(a)}: Real lever with potentiometer and Arduino.\\ 
\textbf{(b)}: Lever model in Gazebo.}
\end{figure}

The lever that is shown in \Cref{fig:lever_real} was used for the experiments. The angle of the lever was measured using a potentiometer and an Arduino Uno. Using the open-source 3D graphics software Blender, a mesh for this lever was also created and is shown in Gazebo in \Cref{fig:lever_gazebo}, and a \gls{urdf} model was created with the mesh as a base. The \gls{urdf}-model requires various properties, such as the specification of the links and joints of a model, visual and collision meshes, dynamics, and friction. These properties were approximated by trial-and-error using the real-world environment and the simulated environment. Both PyBullet and Gazebo can use models of the \gls{urdf} format.

\begin{figure}
\begin{subfigure}{.24\textwidth}
  \centering
  \includegraphics[trim=0 11cm 0 15cm,clip,height=3.3cm,width=\linewidth ]{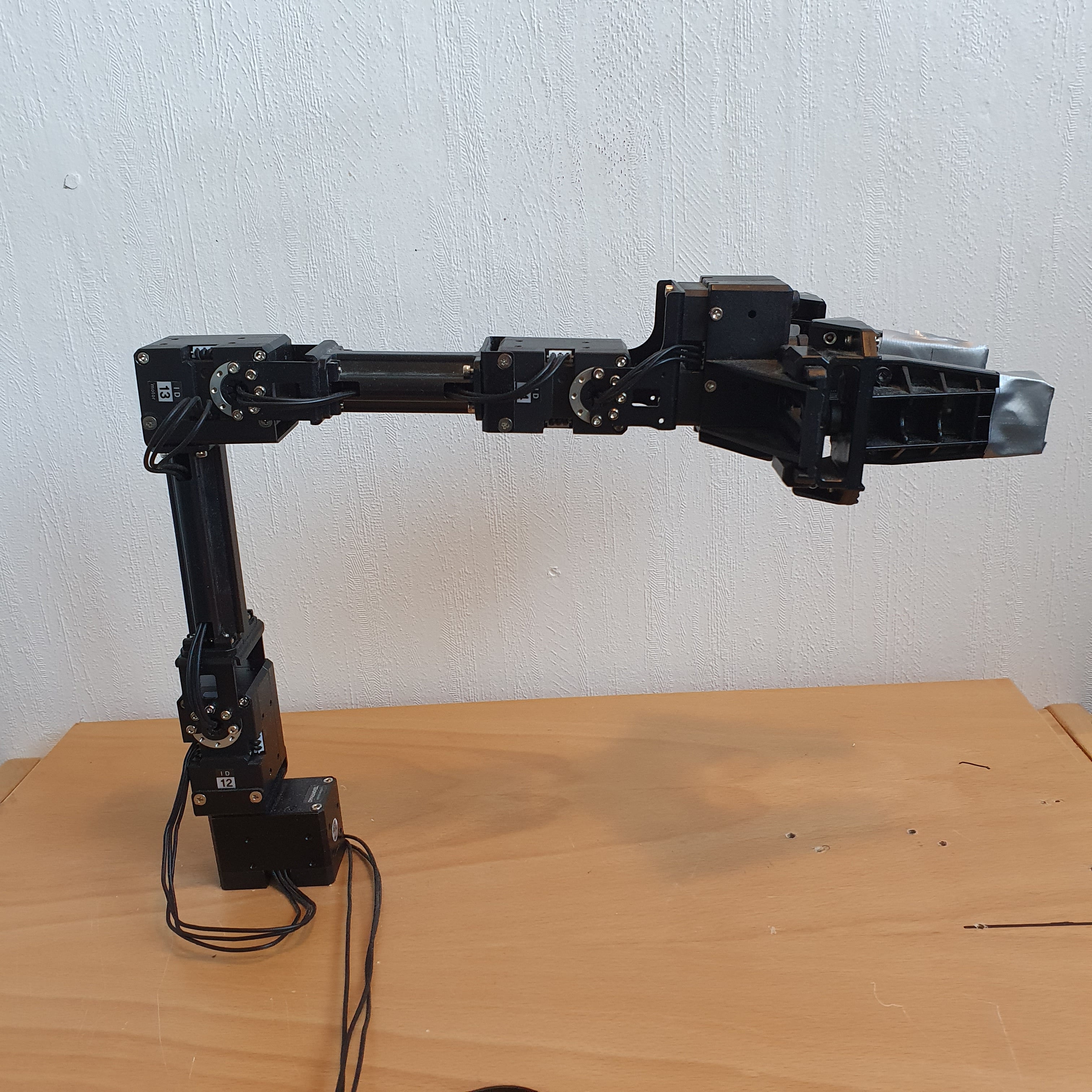}
  \caption{}
  \label{fig:manipulator_real}
\end{subfigure}
\begin{subfigure}{.24\textwidth}
  \centering
  \includegraphics[trim=25cm 10cm 25cm 10cm,clip,width=\linewidth, height=3.3cm]{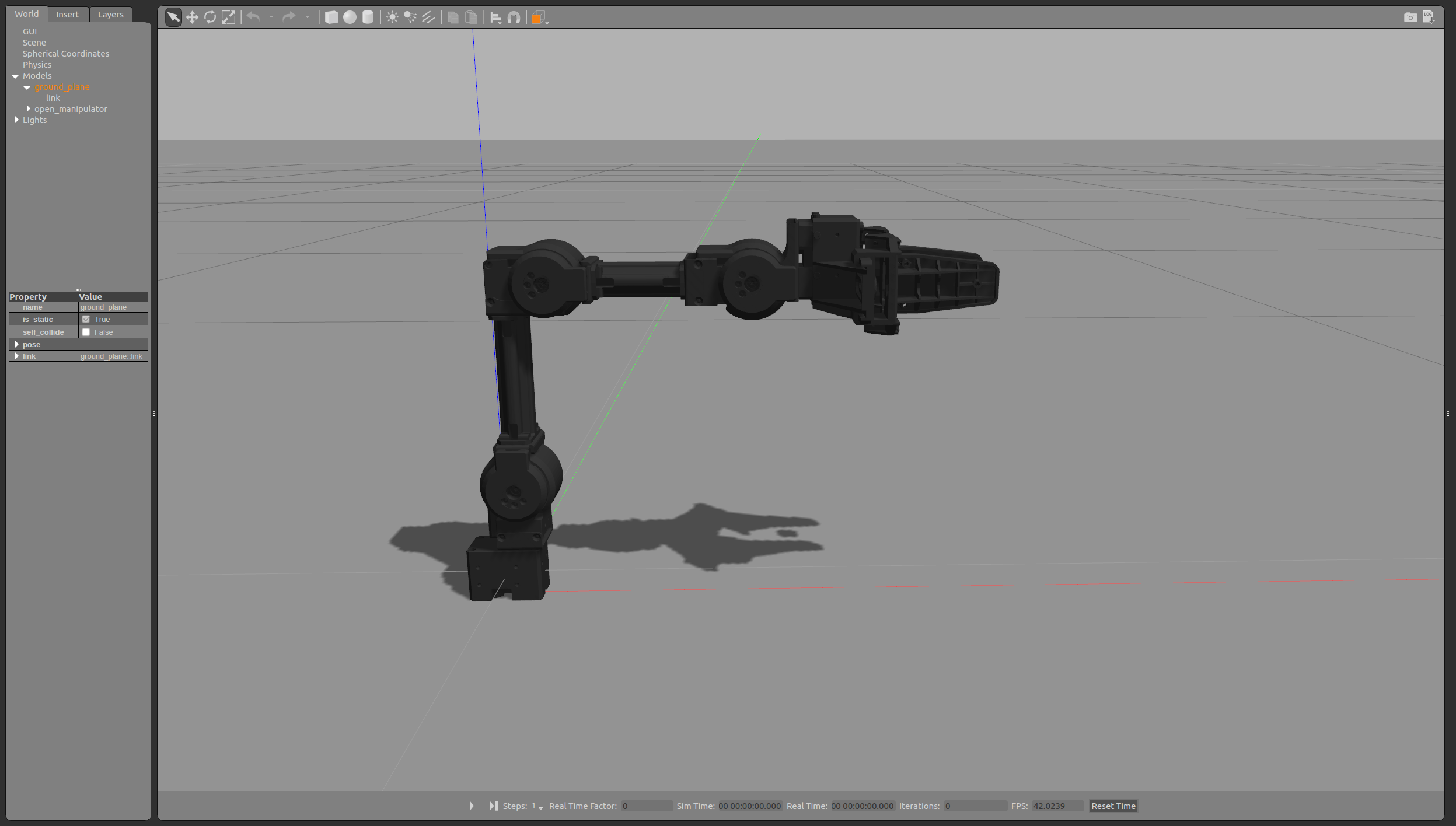}
  \caption{}
  \label{fig:manipulator_gazebo}
\end{subfigure}%
\caption{\textbf{(a)}: Real-world manipulator.\\ 
\textbf{(b)}: Manipulator in Gazebo. 
}
\label{fig:fig}
\end{figure}

The OpenMANIPULATOR-X can be seen in the real-world environment in \Cref{fig:manipulator_real}, and the simulated manipulator can be seen in Gazebo in \Cref{fig:manipulator_gazebo} (the PyBullet version looks similar, and has therefore not been included).

\setlength{\textfloatsep}{2pt}

\subsection{State and Action}
The state-space of this task is of dimension 20. The first 19 entries in the state-space are the observation of the environment. This observation consists of the angles and velocities of the manipulator's joints, the Cartesian position of the lever's base relative to the manipulator's base, the relative distance between the end-effector and the lever's base, and the current angle of the lever. The remaining entry in the state-space is the goal state, and in this task, the goal state is the desired lever angle. When substituting goals with \gls{her}, the desired lever angle is then substituted with the achieved lever angle.


The agent is given a restriction in that it cannot move the manipulator's first joint. This is the joint that rotates around the base of the manipulator. This was done to make training faster, and also because it is trivial to solve for the angle of this joint given the Cartesian x- and y-coordinates of the lever's base: $$\theta_1 = \text{arctan2}(y_{lever},x_{lever}).$$


Since the agent is not able to move the first joint, the action-space is of dimension four. The first three entries correspond to the desired relative angles of joints 2-4, and the fourth entry corresponds to whether the gripper should open or close: 
\begin{equation*}
\begin{split}
a_4 \geq 0, & \rightarrow \text{Gripper should open} \\
a_4 < 0, & \rightarrow \text{Gripper should close}
\end{split}
\end{equation*}
\subsection{Training Procedure}\label{sec:training_procedure}

The training procedure for this task involves to first train the policy in a simulated environment, then transfer the policy to the real environment. The first training takes place in PyBullet, which runs faster on this task compared to Gazebo. PyBullet also has more suitable functionality for \gls{drl}, for instance, a dedicated step function that steps the simulator one time-step forward. The policy was trained in PyBullet for 50 epochs with 30 \textit{training episodes} in each epoch. The policy was then more finely tuned by transfer learning in Gazebo for additional 300 episodes. Gazebo is more accurate concerning the real-world environment on this task. Except for the learning rate, the remaining hyperparameters were the same for Gazebo and PyBullet. The hyperparameters can be seen in \Cref{tab:ddpg_hyperparam}. The action selected by the actor is scaled by a factor of $a_{s}$ before the action is applied to the manipulator. After transfer learning in Gazebo, experiments using the physical manipulator were conducted. 

\begin{table}[]
\vspace{0.2cm}
\centering
\begin{tabular}{@{}ll@{}}
\toprule
Hyperparameter                   & Value  \\ \midrule
Learning rate actor (PyBullet), $\alpha_a$  & 0.001  \\
Learning rate critic (PyBullet), $\alpha_c$ & 0.001  \\
Learning rate actor (Gazebo), $\alpha_a$  & 0.0008  \\
Learning rate critic (Gazebo), $\alpha_c$ & 0.0008  \\
Discount factor, $\gamma$        & 0.98   \\
l2 regularization, $\lambda$         & 1      \\
noise\_eps, $\epsilon_n$         & 0.2    \\
random\_eps, $\epsilon_r$        & 0.18   \\
HER ratio to be replaced, $k$    & 4      \\
HER replay strategy              & future \\
Mini-batch size                  & 256    \\ 
Neurons in input layer, actor    & 20 \\
Neurons in input layer, critic   & 24 \\
Hidden layers, actor and critic  & 3 \\
Neurons in hidden layers, actor and critic         & 256 \\
Neurons in output layer, actor    & 4 \\
Neurons in output layer, critic   & 1 \\
Activation functions hidden layers  & ReLU \\
Activation functions input layer, actor  & Tanh \\
Activation functions input layer, critic  & Linear \\
$a_{s}$  & 0.1 \\ \bottomrule
\end{tabular}
\caption{DDPG hyperparameters}
\label{tab:ddpg_hyperparam}
\end{table}

For this paper, exploration is done using both $\epsilon$-greedy exploration and by adding exploration noise to the action. By $\epsilon$-greedy exploration, with probability $(1-\epsilon_r)$, the agent performs the greedy action, i.e. the best available action according to the actor's knowledge at that time. With probability $\epsilon_r$, the agent performs a random action, hence allowing the agent to explore, with $\epsilon_r\in [0,1]$. For the exploration noise, when the greedy action is selected by the $\epsilon$-greedy exploration, Gaussian noise with mean $\mu = 0$ and standard deviation $\sigma = \epsilon_n$ is added to the greedy action.

During training, the manipulator is randomly selected to start in a grasping position on the lever for half of the episodes. This is done to make the training faster and was inspired by how the authors in \cite{andrychowicz2017hindsight_her} did the pick-and-place task.

As previously mentioned, the goal and start angle of the lever was randomized between episodes. In addition to this, the position of the lever's base relative to the manipulator's base was also randomized between episodes when training in the simulators. In the real-world environment, the lever was attached to a wooden plate together with the manipulator, and could not be moved between episodes. To give the agent the information it needs to calculate the position of the lever when doing the real-world experiments, the distance between the manipulator's base and the lever's base was measured by hand and given as a constant in the state returned from the environment. 

After the training is completed, we run five \textit{test episodes} in both the simulated environments and the physical environment, the results of this will be shown in \Cref{sec:results_discussion}. In these plots, the error measured by the difference in the achieved and the goal lever angle is plotted on the y-axis over the episodes. A video of the real-world experiments is also available\footnote{\url{https://youtu.be/eyPmYoAZqNA    }\vspace{0.05cm}}.

\subsection{Explaining the agent's decisions with SHAP}

The \gls{shap} method was used in an attempt to get a better understanding of the agent's decisions. From the five test episodes, four episodes were used to generate the explanation model, and the remaining episode was explained. The data for the explanations were generated using the real-world manipulator. The \gls{shap} values over the entire first episode are plotted using a force plot for each of the four actions. Features that push the prediction higher or lower are respectively shown in red or blue. The height of each feature contribution shows the magnitude of each feature's contribution.

\section{Results and Discussion}\label{sec:results_discussion}
In this section, we first discuss and compare the performance of the agent on the three different platforms. We discuss reasons why the agent performed more poorly when transferred to the real manipulator. Lastly, we try to get an insight into how the agent makes its decisions by looking at how \gls{shap} assigns importance to the input features for the actions selected.

\begin{figure*}[h]
\vspace{0.2cm}
\caption{Lever angle errors for all episodes and platforms}
\begin{subfigure}{.33\textwidth}
\centering
  \includegraphics[width=1.0\textwidth]{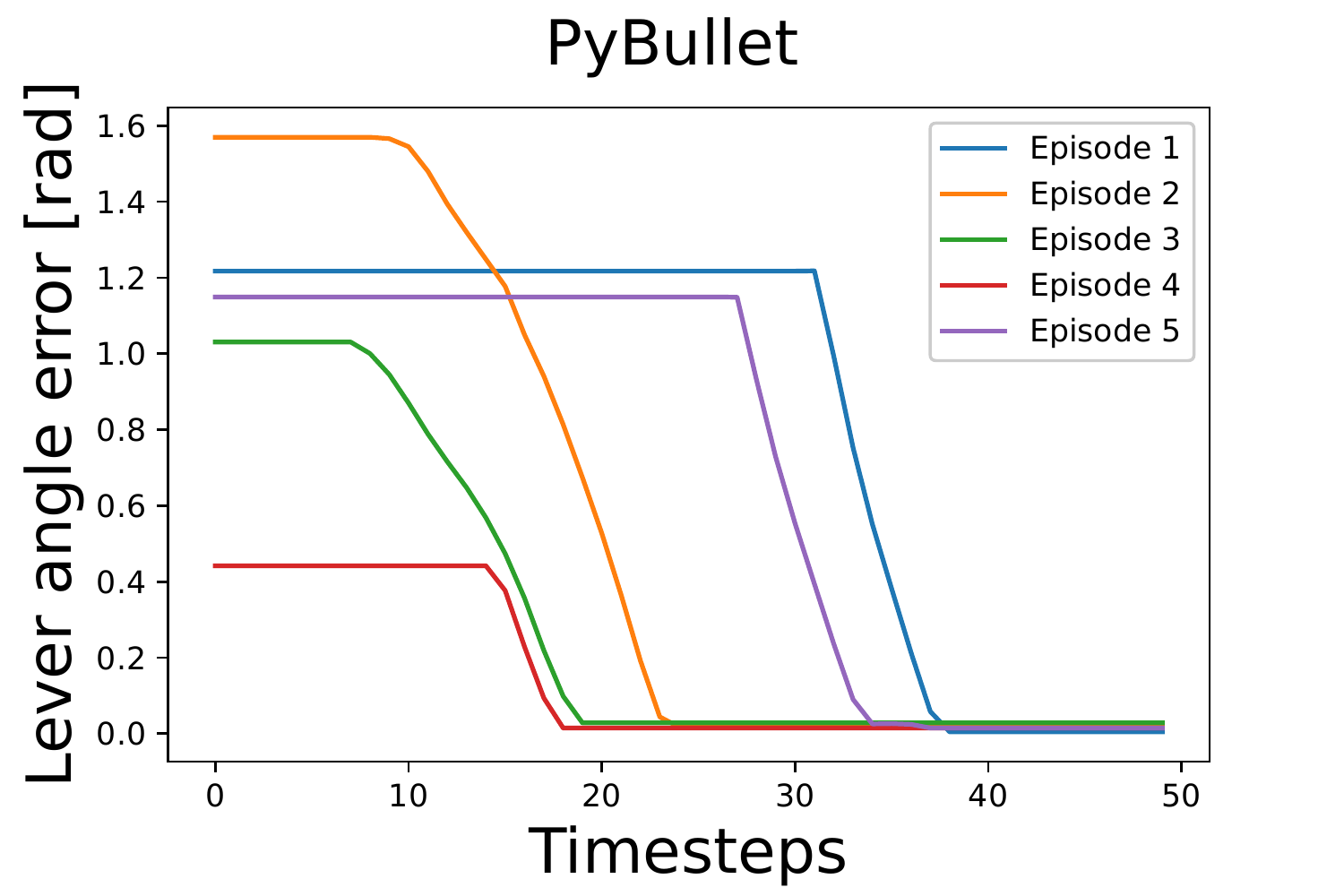}
  \caption{Error, PyBullet}
  \label{fig:error_lever_manipulation_pybullet}
\end{subfigure}
\begin{subfigure}{.33\textwidth}
  \flushright
  \includegraphics[width=1.0\textwidth]{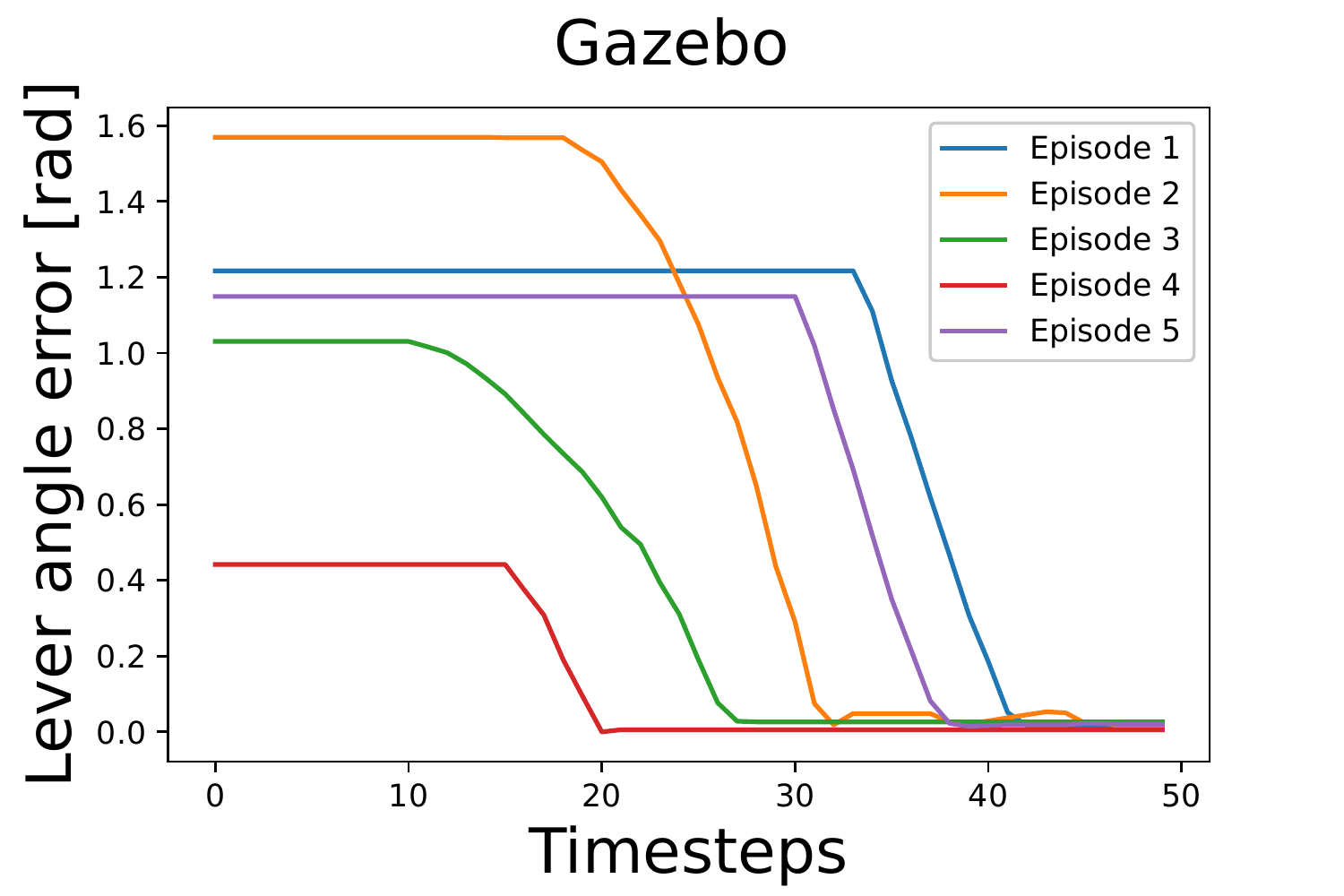}
  \caption{Error, Gazebo after transfer learning}
  \label{fig:error_lever_manipulation_gazebo_with_tf}
\end{subfigure}
\begin{subfigure}{.33\textwidth}
  \flushright
  \includegraphics[width=1.0\textwidth]{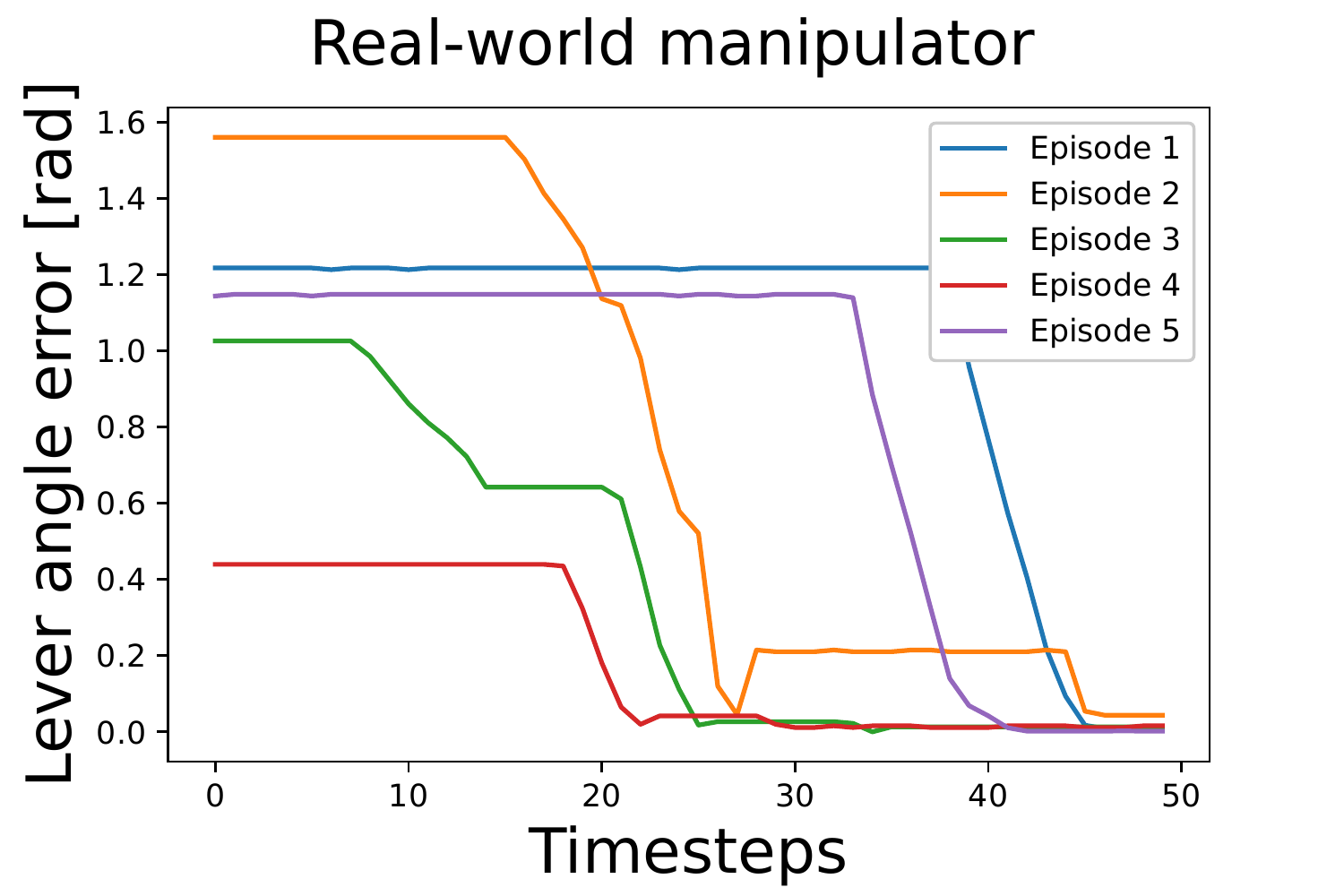}
  \caption{Error, real world manipulator}
  \label{fig:error_lever_manipulation_real_world}
\end{subfigure}
\label{fig:shap_lmt_expl}

\vspace{-1cm}
\end{figure*}

\subsection{Lever manipulation}

\begin{table}[]
\vspace{0.2cm}
\begin{tabular}{@{}llll@{}}
\toprule
Episode & $|\theta_{start}-\theta_{goal}|$ & Initial distance from end-effector to lever \\ \midrule
1       & $1.21719255\text{ }\si{\radian}$                       & $0.2674\si{\meter}$                                  \\
2       & $1.56957698\text{ }\si{\radian}$                       & $0.1763\si{\meter}$                                   \\
3       & $1.03069065\text{ }\si{\radian}$                       & $0.1742\si{\meter}$                                   \\
4       & $0.44181838\text{ }\si{\radian}$                       & $0.1833\si{\meter}$                                   \\
5       & $1.14932491\text{ }\si{\radian}$                       & $0.2508\si{\meter}$                                   \\ \bottomrule
\end{tabular}
\caption{Initial situations for the experiments}
\label{tab:initial_conditions}
\end{table}

\begin{figure}
    \centering
    \includegraphics[width=1.0\linewidth]{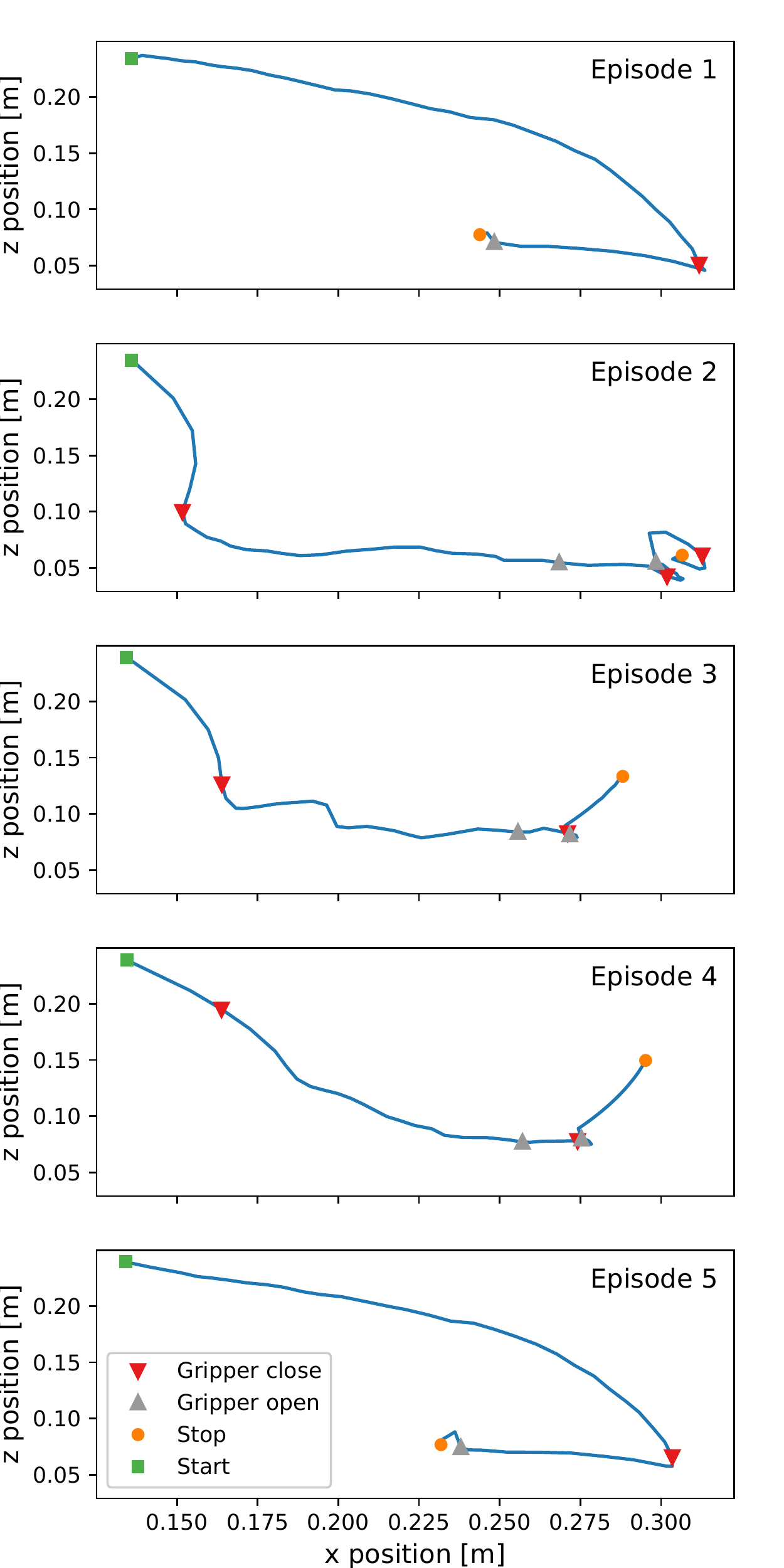}
    \caption{Task space trajectories for the real-world manipulator, with indications for changes in gripper state. The y-coordinate is constant since the first joint is not moved.}
    \label{fig:task_space_gripper}
\end{figure}

The results from the lever manipulation experiments can be seen in \crefrange{fig:error_lever_manipulation_pybullet}{fig:error_lever_manipulation_real_world}, where the results were achieved by randomly selecting five different start and goal conditions, and running these over five episodes for both the simulated and physical platforms. First we will discuss the results from the simulated environments in \Cref{fig:error_lever_manipulation_pybullet} and \Cref{fig:error_lever_manipulation_gazebo_with_tf}. These results are quite good and show smooth trajectories for all episodes. Some episodes take more time to complete than others, for instance, Episode 1 and Episode 5. From \Cref{tab:initial_conditions} we can see that these two episodes are when the lever initiates furthest away from the manipulator, and the initial lever angle error is also among the largest, so naturally, these episodes are the most time-consuming. The performance in both simulated environments is also quite similar, with Gazebo being somewhat slower for all episodes. The operation in Gazebo could be faster if more episodes of transfer learning were done.

We will now discuss the differences between the physical environment and simulated environments. From the plots of Episode 2 and Episode 3 in \Cref{fig:error_lever_manipulation_real_world}, which show the performance in the physical environment, it is noticeable that the performance is not as good as in the simulated environments. This can also be seen in \Cref{fig:task_space_gripper}, especially in Episode 2, where the trajectory in the bottom right corner seems chaotic. This discrepancy in the real-world results likely stems from modeling errors in the simulated environments compared to the real-world environment. This phenomenon is described in \cite{RobReinforceSurvey} as under-modeling. This under-modeling is especially noticeable in Episode 2, where the manipulator first moves the lever too far, and then when it tries to correct this, it first tries to grasp too low on the lever and instead grasps the base of the lever. In Episode 3 and Episode 4, the manipulator also first moves the lever too far. This could indicate that both the dimensions of the lever and the friction of the lever could be under-modeled in the simulators. If more effort was made to more accurately model the friction and dimensions of the lever, these problems could have been avoided. However, friction is notoriously hard to model, and it is also possible that the effort to make the simulated lever more similar to the real lever, would fail. It is important to note that the agent was not trained at all on the real manipulator. If safe training on the physical manipulator could be implemented, this would likely increase the performance by enabling the agent to learn how to act according to the real environment, and not according to the simulation.

A possible way to improve the results from the real-world experiments could be to use \emph{Dynamics randomization}. This technique was proposed in \cite{peng2017simtoreal} and involves using a \gls{rnn} to enable the agent to approximate the dynamics of the system while randomizing the dynamics of the simulated system between episodes. This technique could, for instance, be applied so that the agent can approximate the dynamics of the real-world lever.

As can be seen in the literature, the advantages and disadvantages of \gls{drl} compared to more traditional robotic control methods are many. However, the main disadvantage of robotic \gls{drl} is arguably the lack of explanation in the agent's decisions and the general trustworthiness of the agent. This disadvantage was attempted to be remedied by using \gls{shap} in this paper.


\begin{figure}
    \vspace{0.2cm}
    \centering
    \includegraphics[width=1.0\linewidth]{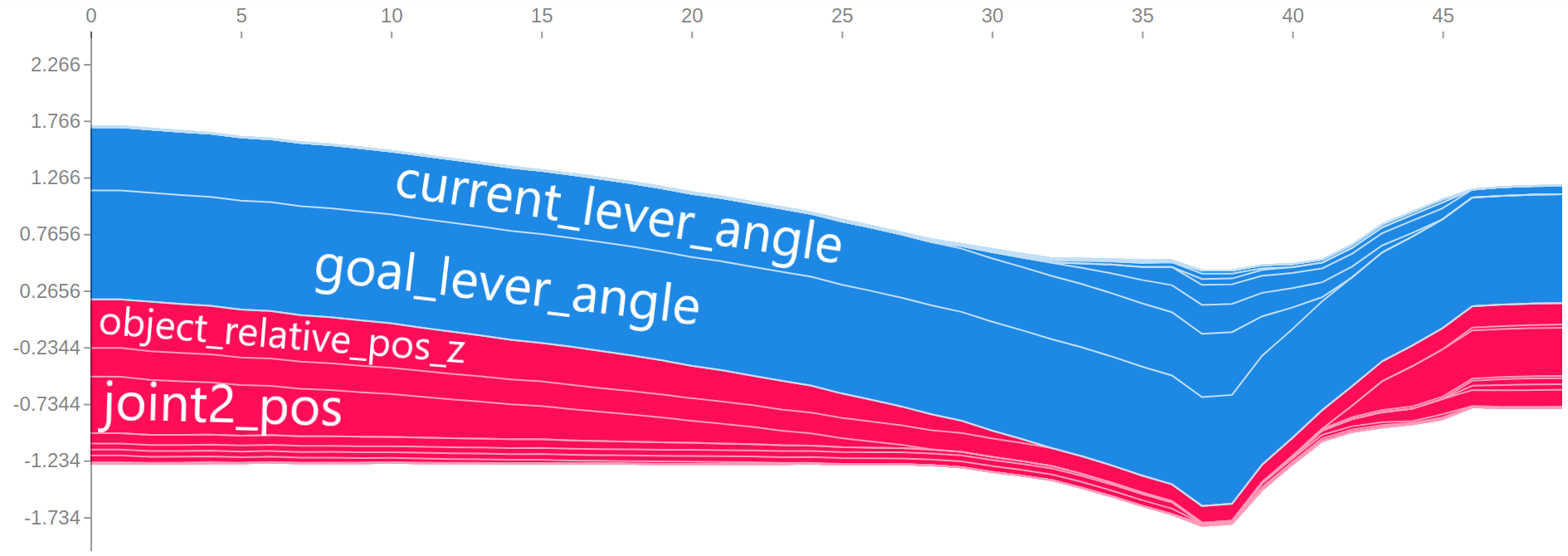}
    \caption{SHAP values for action $a_1$ in Episode 1}
    \label{fig:action1_shap_episode1}
    \vspace*{\floatsep}
    
    \centering
    \includegraphics[width=1.0\linewidth]{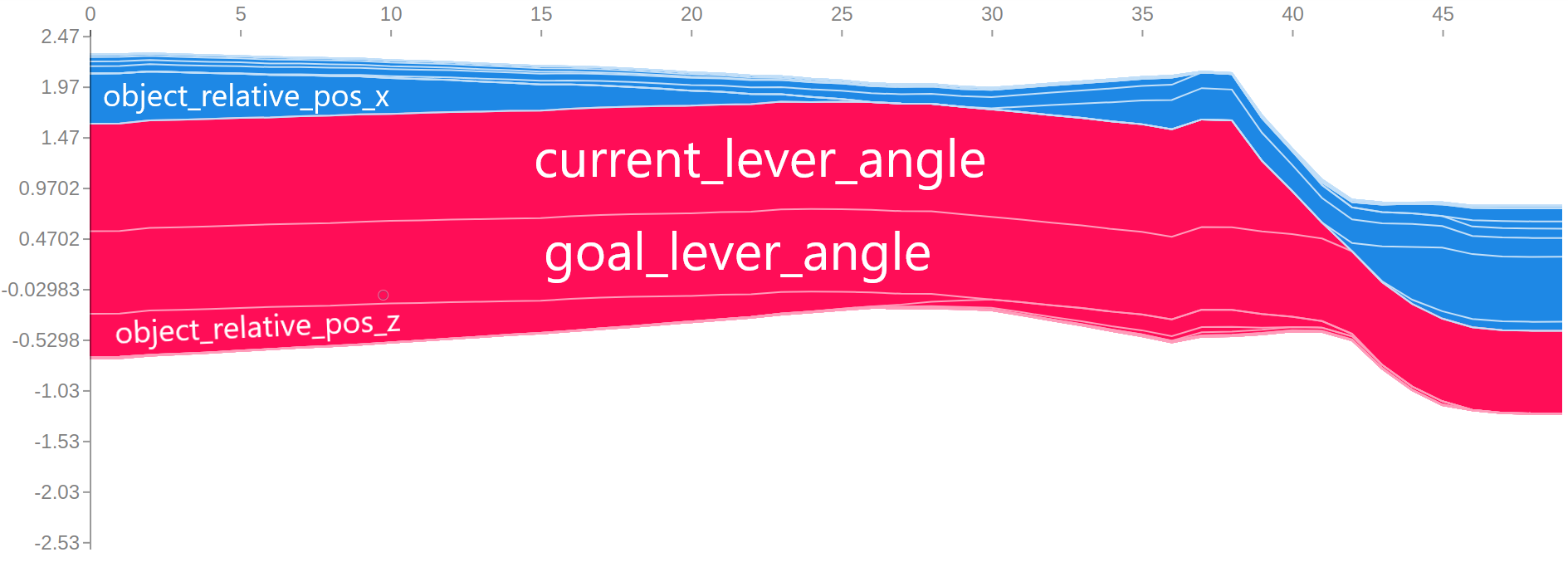}
    \caption{SHAP values for action $a_2$ in Episode 1}
    \label{fig:action2_shap_episode1}
    \vspace*{\floatsep}
    
    \centering
    \includegraphics[width=1.0\linewidth]{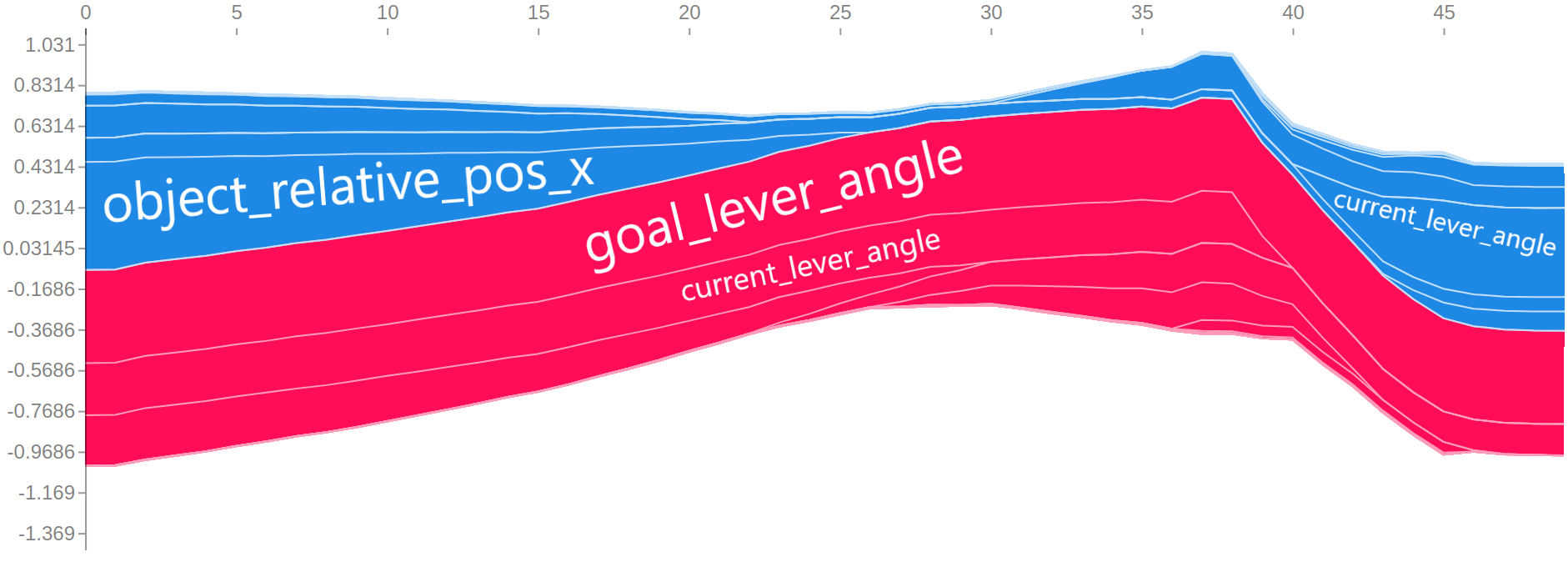}
    \caption{SHAP values for action $a_3$ in Episode 1}
    \label{fig:action3_shap_episode1}
    \vspace*{\floatsep}
    
    \centering
    \includegraphics[width=1.0\linewidth]{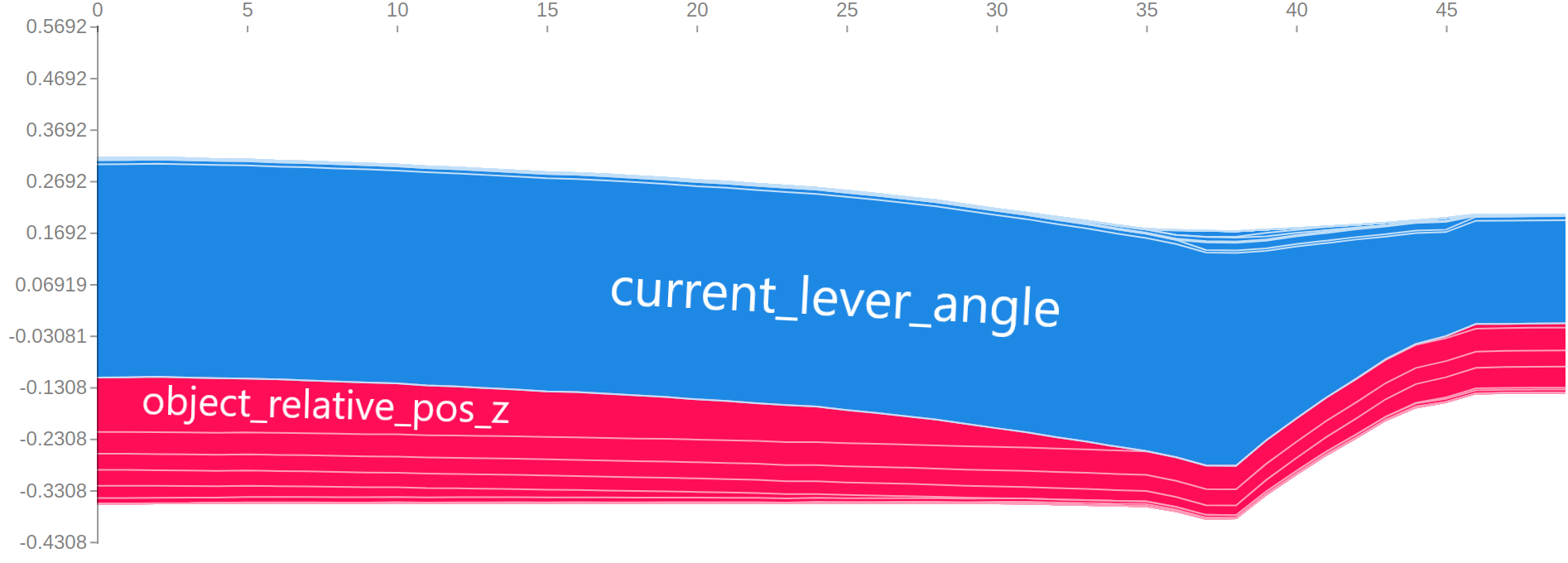}
    \caption{SHAP values for action $a_4$ in Episode 1}
    \label{fig:action4_shap_episode1}
\end{figure}

\subsection{SHAP explanations}

The \gls{shap} values plotted over the operation in Episode 1 can be seen in \crefrange{fig:action1_shap_episode1}{fig:action4_shap_episode1}. In the plots, the x-axis corresponds to the step numbers on the x-axis of the plot in \Cref{fig:error_lever_manipulation_real_world}. According to these \gls{shap} values, the most important states for the agent to make its decisions are the current lever angle and the desired lever angle. This conforms with intuition since, without these two states, it would be virtually impossible for the agent to move the lever to the desired lever angle. Other important states according to \gls{shap} are the relative distance on the x- and z-axis between the end-effector and the lever's base. This also makes sense, since these two states, combined with the current lever angle, can be used to calculate where the end-effector is relative to the lever. However, it is surprising that the manipulator's joint angles contribute so little to the actions. Intuitively, the angles of the manipulator's joints should be important for knowing which action to execute, considering that actions 1-3 correspond to how these joints should be moved. It is also possible that the validity of these \gls{shap} values might be put into question. \gls{shap} assumes that all states are independent, which they might not be in most robotics applications \cite{kumar_problems_2020}.

Other than which states are important for the decisions of the agent, it is difficult to interpret these plots in a way that either increases or decreases the trust in the agent's decisions. An interesting point was made by \cite{kumar_problems_2020} regarding whether \gls{shap} contributes to understanding whether or not a decision is correct. They state that humans often explain by using contrastive statements (e.g. why A rather than B). It might be difficult to use a feature attribution method for this since they only describe why and not why not. It might be that for such complex problems as robotic manipulation, it is not that helpful to use a single tool to explain all problems. Instead, it might be beneficial to attempt to tailor the explanation mechanism to the use-case \cite{kumar_problems_2020}. The creators of \gls{lime} say in \cite{ribeiro2016whyshoulditrustyou_lime}, that for an explanation to be easy to understand, the features themselves should be easy to understand. They also say that the features used by the \gls{ml} model and inputs used by the explanation model need not necessarily be the same if this can make the explanations more understandable. It is possible that the quality of the explanations could be increased if the features were transformed into an input that is more easily understood (e.g. by going from joint space to task space).

\gls{drl} research often seems to have the goal to either increase the performance of the agents or to make algorithms that can solve even more complex environments. Better performance is arguably important, but as long as the decisions of the agents cannot be explained, \gls{drl} will likely remain a tool for solving toy problems, and will not be used in safety-critical applications. This is why increasing the understanding of the agents, for instance, through \gls{xai} methods might be necessary. The implementations in this paper aim at demonstrating how well a state-of-the-art \gls{xai} method works on a robotic manipulator, and for identifying the right questions to come up with improved methods that will help make \gls{drl} safer in such applications.

\section{Conclusion}
In this paper, we have shown how a Deep Reinforcement Learning policy can be implemented on a real-world robotic manipulator using \gls{ros} and used to manipulate objects, in this case, a lever. Hindsight Experience Replay simplifies the process of designing a reward function and is useful for a complex system such as a robotic manipulator. The agent performs very well when controlling the simulated manipulator, and performs comparably for most episodes when controlling the real-world manipulator. As stated in the discussion, the differences between the performance in the simulated and real environments likely stems from under-modeling of the lever in the simulations.

To explain the agent's decisions, we implemented the Explainable Artificial Intelligence method \gls{shap}, which gives results that mostly agree with human intuition. The two most important states, according to \gls{shap}, are the current lever angle and the goal lever angle, which is reasonable. However, it is unexpected that some states that seem like should be important, such as the joint variables, are not very important according to \gls{shap}. This divergence from human intuition may stem from correlations between the states in the system, which may make the Shapley value approximation less accurate.


\section*{Acknowledgment}
This work was supported by the Research Council of Norway through the EXAIGON project, project number 304843.

\bibliographystyle{IEEEtran}
\bibliography{mylib}

\end{document}